\documentclass[10pt,twocolumn,letterpaper]{article}

\usepackage{iccv}
\usepackage{times}
\usepackage{epsfig}
\usepackage{graphicx}
\usepackage{amsmath}
\usepackage{amssymb}
\usepackage{ulem}
\usepackage{array}
\usepackage{multirow}
\usepackage{bm}
\usepackage{subfigure}

\usepackage{colortbl}

\usepackage[breaklinks=true,bookmarks=false]{hyperref}

\iccvfinalcopy % *** Uncomment this line for the final submission

\def\iccvPaperID{9066} % *** Enter the ICCV Paper ID here

\ificcvfinal\fi

\begin{document}

%%%%%%%%% TITLE
\title{PR-GCN: A Deep Graph Convolutional Network with Point Refinement \\for 6D Pose Estimation}
\author{Guangyuan Zhou$^{1}$, Huiqun Wang$^{1, 2}$, Jiaxin Chen$^{2}$ and Di Huang$^{1, 2}$ \thanks{indicates the corresponding author.}\\
$^{1}$State Key Laboratory of Software Development Environment, Beihang University, Beijing, China 
\\
$^{2}$School of Computer Science and Engineering, Beihang University, Beijing, China
\\
{\tt\small \{zhouguangyuan, hqwangscse, jiaxinchen, dhuang\}@buaa.edu.cn}}

%\and

\maketitle
\ificcvfinal\thispagestyle{empty}\fi

%%%%%%%%% ABSTRACT
\begin{abstract}
RGB-D based 6D pose estimation has recently achieved remarkable progress, but still suffers from two major limitations: (1) ineffective representation of depth data and (2) insufficient integration of different modalities. This paper proposes a novel deep learning approach, namely Graph Convolutional Network with Point Refinement (PR-GCN), to simultaneously address the issues above in a unified way. It first introduces the Point Refinement Network (PRN) to polish 3D point clouds, recovering missing parts with noise removed. Subsequently, the Multi-Modal Fusion Graph Convolutional Network (MMF-GCN) is presented to strengthen RGB-D combination, which captures geometry-aware inter-modality correlation through local information propagation in the graph convolutional network. Extensive experiments are conducted on three widely used benchmarks, and state-of-the-art performance is reached. Besides, it is also shown that the proposed PRN and MMF-GCN modules are well generalized to other frameworks.
\end{abstract}

%%%%%%%%% BODY TEXT
\section{Introduction}
% problem statement
6D pose estimation aims to predict the orientation and location of an object in the 3D space from a canonical frame. It has received extensive attention in computer vision, since it is one of the fundamental steps for a wide range of applications, such as robotics grasping \cite{ColletMS11,realtime,voxelnet} and augmented reality \cite{DBLP:conf/eccv/ManhardtKNT18,DBLP:journals/tvcg/MarchandUS16}. Traditional methods \cite{DBLP:conf/iccv/HinterstoisserHCIKNL11, linemod} attempt to accomplish this task based on RGB images only. They adopt handcraft features (\eg SIFT \cite{Lowe99} and SURF \cite{surf}) to establish correspondence between input and canonical images. Inspired by the great success in detection/recognition, deep neural networks are recently explored to address this issue, including the single-stage regression methods \cite{posenet} and key-point based methods \cite{segmentation, bb8, viewsandkeypoints, realtime, DBLP:conf/eccv/OberwegerRL18, DBLP:conf/eccv/NewellYD16,DBLP:conf/cvpr/LiuJAK20}. Despite the remarkable promotion in accuracy, RGB-based deep models heavily rely on textures; thus sensitive to illumination variations, severe occlusions, and cluttered backgrounds.

\begin{figure}[!t]
	\centering
	\includegraphics[width=0.99\linewidth]{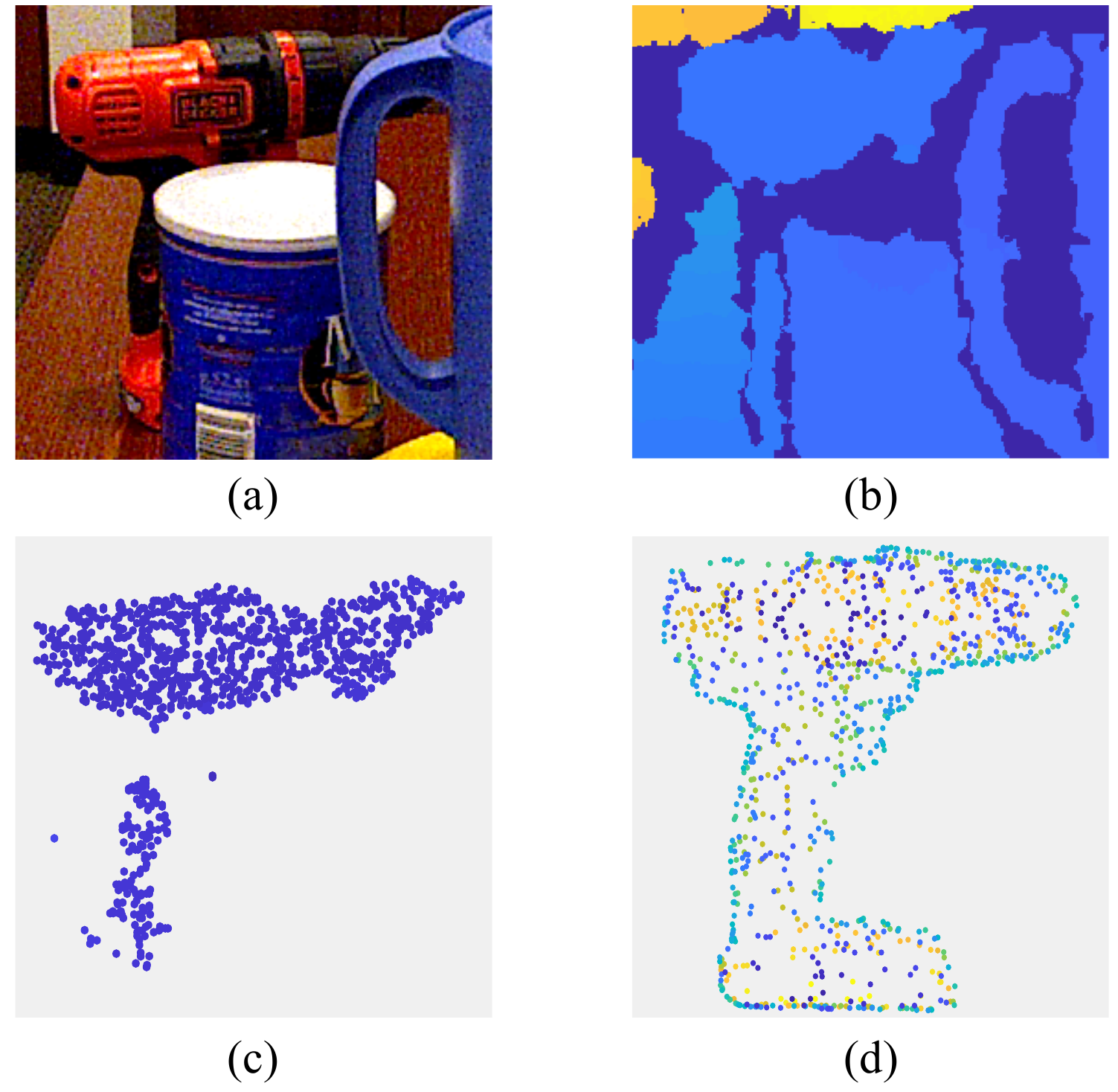}
	\caption{Example object: (a) / (b) are RGB / depth images; (c)/(d) are
	generated incomplete noisy point cloud and ground-truth.}
	\label{fig:issues}
	%\vspace{-0.4cm}
\end{figure}

Along with the emergence and innovation of depth sensors, 6D pose estimation on RGB-D data has become popular, expecting to deliver performance gain by adding geometry information. Early works \cite{linemod, Michel, zeng} estimate object poses from RGB images and refine them according to depth maps. Later studies \cite{densefusion, MCN} dedicate to integrating RGB and depth clues in a more sophisticated way. Particularly, \cite{pointfusion,frustrum,votenet,densefusion,pvn3d} represent depth images as 3D point clouds and the models are more efficient in computation and storage than those on original depth maps. By jointly making use of both modalities, RGB-D based solutions report better scores, with the superiority in the presence of the difficulties aforementioned as well as the low-texture case. 

However, current RGB-D pose estimation suffers from two major limitations: \textbf{ineffective representation of depth data} and \textbf{insufficient combination of two modalities}. For the former, as captured in cluttered scenes, depth information is usually noisy and incomplete (see Fig. \ref{fig:issues}). Inferring poses from such data, either in 2D depth maps or 3D point clouds, is not robust, leading to accuracy deterioration. For the latter, RGB and depth clues are fused by concatenating separately learned single-modal features \cite{densefusion} or by applying a simple point-wise encoder \cite{pvn3d}, where inter-modality correlations are not considered or roughly modeled in a global manner, leaving much room for improvement. 

\begin{figure*}[!t]
	\centering
	\includegraphics[width=1.0\linewidth]{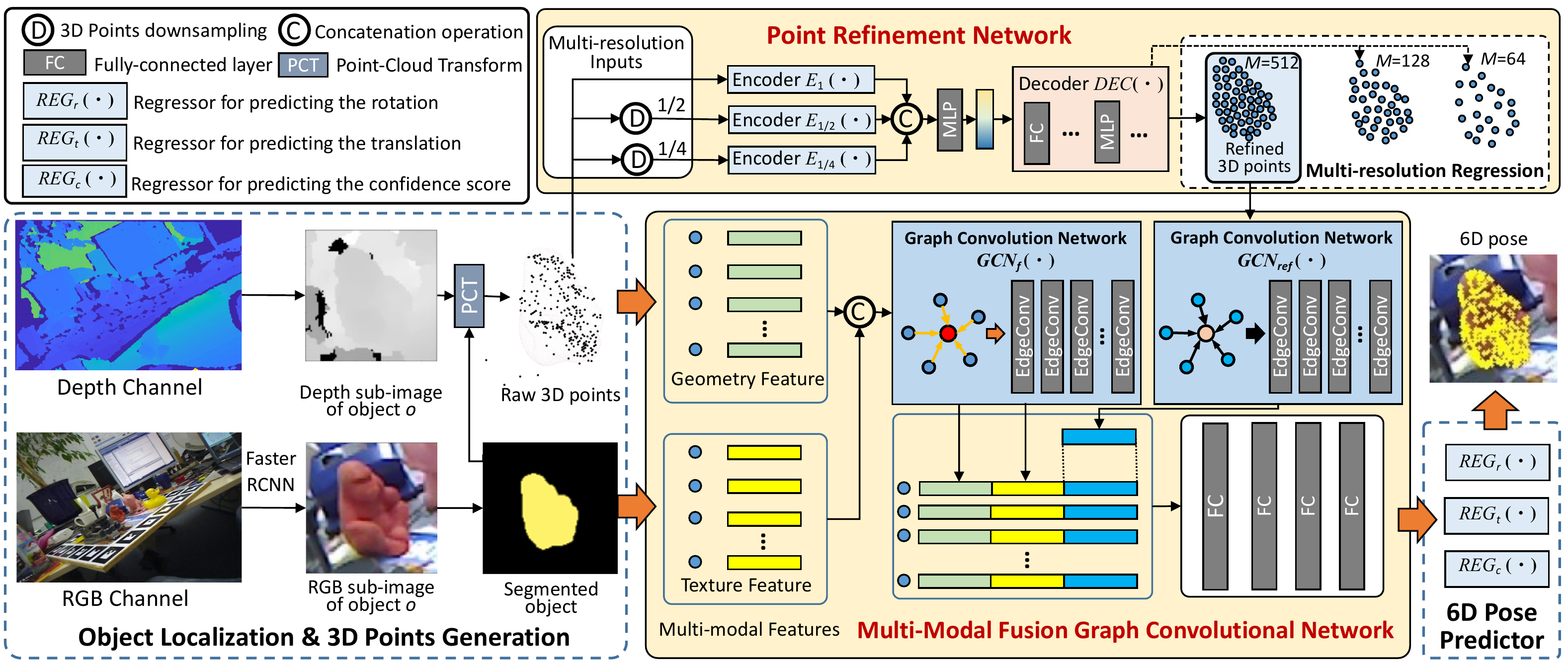}
	\caption{Illustration of PR-GCN. Given an RGB-D image, it first localizes objects on RGB images and generates their raw 3D point clouds. Subsequently, PRN generates refined 3D points to polish shape clues and MMF-GCN integrates multi-modal features by propagating local geometry-aware information and leveraging refined 3D points. 6D pose is finally inferred based on the feature delivered by MMF-GCN.}
	\label{fig:pipeline}
	\vspace{-0.4cm}
\end{figure*}

In this paper, we propose a novel deep learning approach, namely Graph Convolutional Network with Point Refinement (PR-GCN), to simultaneously address the two limitations in a unified way. As in Fig.~\ref{fig:pipeline}, given the RGB image and 3D point cloud (generated from depth map) of an object, we first introduce a Point Refinement Network (PRN) to polish the point cloud. Endowed with an encoder-decoder structure and trained with a regularized multi-resolution regression loss, PRN recovers the missing parts of the raw input with noise removed. Subsequently, we integrate RGB-D clues by a Mutli-Modal Fusion Graph Convolutional Network (MMF-GCN). It constructs a $k$-Nearest Neighbor ($k$-NN) graph and extracts geometry-aware inter-modality correlation through local information propagation in the Graph Convolutional Network (GCN). An additional $k$-NN graph and GCN are employed to encode local geometry attributes of the refined point cloud as a complement to the original data. The features from the two GCNs are then combined and fed into several fully-connected layers for final 6D pose prediction. We extensively evaluate PR-GCN on three public benchmarks, Linemod \cite{linemod}, Occlusion Linemod \cite{brachmann2014learning}, and YCB-Video \cite{POSECNN}, and achieve the state-of-the-art performance. We also show that the proposed PRN and MMF-GCN modules are well generalized to other frameworks.

The contributions: 1) We propose the PR-GCN approach to 6D pose estimation by enhancing depth representation and multi-modal combination. 2) We present the PRN module with a regularized multi-resolution regression loss for point-cloud refinement. To the best of our knowledge, it is the first that applies 3D point generation to this task. 3) We develop the MMF-GCN module to capture local geometry-aware inter-modality correlation for RGB-D fusion.

%-------------------------------------------------------------------------
\section{Related Work}

\noindent \textbf{RGB based 6D Pose Estimation.} The traditional methods \cite{linemod, DBLP:conf/eccv/GuR10, DBLP:journals/ftcgv/LepetitF05} establish correspondence between object appearances and poses from single RGB images. Linemod \cite{linemod} predicts poses by modeling the relationship between texture gradients and surface normals on 3D templates. \cite{3dcoordinates} exploits key-points of specific objects for pose estimation by iteratively matching them between input and canonical frames. As in other vision tasks, deep models are also investigated to build more powerful features for this issue. DeepIM \cite{deepim} adopts CNNs to learn reliable representations for template-matching. BB8 \cite{bb8} applies CNNs in a multi-stage segmentation scheme to regress key-point  coordinates. PVNet \cite{pvnet} proposes a deep offset prediction model to alleviate negative impacts of occlusions. CDPN \cite{cpdn} and Pix2pose \cite{pix2pose} map 3D coordinates to 2D pixels and regress pose parameters on 2D images. LatentFusion \cite{park2020latentfusion} handles unseen object poses by reconstructing a latent 3D representation.

\noindent \textbf{RGB-D based 6D Pose Estimation.} With geometry information, depth maps contribute to pose estimation for various lighting conditions and low-textured appearances, complementary to RGB images. MCN \cite{MCN} employs two CNNs for representation learning in RGB and depth respectively and resulting features are then concatenated for pose prediction. PoseCNN \cite{POSECNN} and SSD-6D \cite{SSD-6D} follow the coarse-to-fine scheme, where poses are initially estimated on RGB frames and subsequently refined on depth maps. \cite{MoreFusion} builds a multi-view model to jointly reconstruct whole scenes and optimize multi-object poses.

Recently, there has emerged a trend to represent geometry clues in 3D point clouds rather than depth mas for higher efficiency \cite{densefusion,CF,G2L,frustrum,pvn3d}. DenseFusion \cite{densefusion} designs a heterogeneous network to integrate texture and shape features and such representation proves more discriminative than single-modal ones. CF \cite{CF} introduces attention modules to combine the two modalities for further improvements. G2L \cite{G2L} segments point clouds of objects in scenes by frustum pointnet \cite{frustrum} and regresses pose parameters via extra coordinate constraints. PVN3D \cite{pvn3d} incorporates DenseFusion into 3D key-point detection and instance semantic segmentation, significantly boosting the performance.

Unfortunately, the point clouds generated from the depth maps are often of a low quality, since the shape information is often incomplete and noisy as Fig.~\ref{fig:issues} shows. Besides, the combination of RGB and depth clues is launched in a very rough way, \eg, directly concatenating or point-wise encoding. In contrast, our approach develops the PRN and MMF-GCN modules to polish depth clues by generating refined point clouds and enhance integration by capturing local geometry-aware inter-modality correlations respectively, both of which are beneficial to pose estimation.

%-------------------------------------------------------------------------

\section{The Proposed Method}
\subsection{Framework Overview}
RGB-D based 6D pose estimation recovers 6D poses of objects in RGB-D images, where 6D pose is usually represented by a rotation matrix $\bm{R}\in SO(3)$ and a translation vector $\bm{t}\in \mathbb{R}^{3}$. For this issue, we propose the PR-GCN approach. As Fig.~\ref{fig:pipeline} depicts, it consists of four steps: object localization and 3D points generation, 3D points refinement, GCN-based multi-modal fusion, and 6D pose prediction. 

%3D rotation belongs to rigid body transformation can be represented as $ R = (a, b, c), a,b,c\in R^3, (a \times b) \cdot c = (b \times c) \cdot a = (c \times a) \cdot b = 1$. 

%To address this problem, in this paper we firstly represent the depth information of particular objects into sets of 3D points, and propose a deep graph convolution network with point refinement, which can simultaneously deal with incomplete noisy 3D points and effectively fuse the multi-modal data, \ie, the RGB image and the 3D points. 

%method.  firstly represent the depth information of particular objects into sets of 3D points, and propose a deep graph convolution network with point refinement, which can simultaneously deal with incomplete noisy 3D points and effectively fuse the multi-modal data, \ie, the RGB image and the 3D points. 

\textbf{Object Localization and 3D Points Generation.} Given an RGB-D image $I=(I_{rgb}, I_{d})$, we firstly locate objects on $I_{rgb}$ using the off-the-shelf Faster R-CNN \cite{faster-rcnn} detector, where $I_{rgb}$ and $I_{d}$ denote the RGB and depth channels of $I$. According to the detected bounding boxes, we crop the sub-images $\{I_{o}\}=\{(I_{o,rgb},I_{o,d})\}$, each of which contains an instance $o$. $I_{o,rgb}$ and $I_{o,d}$ are the RGB and depth channels of $I_{o}$. As in PoseCNN \cite{POSECNN}, we add a segmentation head to remove background of $I_{o,rgb}$. With $M_{o}$ (foreground mask) and $I_{o,d}$, raw 3D points $\bm{P}_{o}=[\bm{p}^{(1)}_{o}; \cdots; \bm{p}^{(i)}_{o};  \cdots; \bm{p}^{(N)}_{o}]\in \mathbb{R}^{N\times 3}$ are rendered by Point-Cloud Transform (PCT) \cite{DBLP:books/daglib/0015576}, where $N$ is the number of points and $\bm{p}^{(i)}_{o}\in \mathbb{R}^{3}$ is the 3D coordinate of the $i$-th point. It is worth noting that $\bm{P}_{o}$ could be severely incomplete and noisy due to external occlusions and sensor noise (see Fig. \ref{fig:issues}). 

%In order to remove the cluttered background and perform object localization more precisely, PSPNet \cite{pspnet} is further employed for semantic segmentation on $\{I_{o,rgb}\}$. 

\textbf{3D Points Refinement.} To polish the quality of the generated raw 3D points $\bm{P}_{o}$, we propose the PRN module. As in Fig.~\ref{fig:pipeline}, it is composed of an MLP-based encoder and a multi-resolution decoder to recover the complete and accurate 3D point cloud $\hat{\bm{P}}_{o}=[\hat{\bm{p}}^{(1)}_{o}; \cdots; \hat{\bm{p}}^{(m)}_{o}; \cdots; \hat{\bm{p}}^{(M)}_{o}]\in \mathbb{R}^{M\times 3}$, where $M$ is the number of refined points. A regularized multi-resolution regression loss is formulated in training, enhancing its ability of filtering out noise in $\bm{P}_{o}$. 

\textbf{GCN-based Multi-Modal Fusion.} For more sufficient RGB-D fusion, we propose the MMF-GCN module. As in Fig.~\ref{fig:pipeline}, it extracts texture and geometry features from $I_{o,rgb}$ and $\bm{P}_{o}$, respectively, and a graph is built based on geometry distribution. Accordingly, $I_{o,rgb}$ and $\bm{P}_{o}$ are initially integrated by applying a GCN $GCN_{f}(\cdot)$ on the previously built graph through local information propagation. The geometry clues from the refined 3D points are encoded by introducing an extra GCN $GCN_{ref}(\cdot)$ and then incorporated into the initially fused features, which are fed into several stacked fully-connected layers $T(\cdot)$ for further fusion. The resulting feature $\bm{G}_{o}=[\bm{g}_{o}^{(k)}]_{k=1,\cdots,K}\in \mathbb{R}^{K\times d}$ is therefore the multi-modal representation for successive 6D pose estimation, where $K$ and $d$ refer to the number of points and the feature dimension, respectively. Since MMF-GCN captures local geometry-aware inter-modality correlation and leverages refined 3D point clouds, it is expected to deliver more discriminative and robust features.

\textbf{6D Pose Prediction.} $[\bm{g}_{o}^{(k)}]_{k=1,\cdots,K}$ is finally fed into three regression branches: $REG_{r}(\cdot)$, $REG_{t}(\cdot)$, $REG_{c}(\cdot)$, for rotations $\{\hat{\bm{R}}_{o}^{(k)}=REG_{r}(\bm{g}_{o}^{(k)})\}$, translations $\{\hat{\bm{t}}_{o}^{(k)}=REG_{t}(\bm{g}_{o}^{(k)})\}$, confidence scores $\{s_{o}^{(k)}=REG_{c}(\bm{g}_{o}^{(k)})\}$, respectively. Each branch has four fully-connected layers. Similar to \cite{pvn3d, pvnet, densefusion}, we select the candidate with the highest confidence score as the estimated pose, formulated as:
\begin{equation}\label{eq:pred}
(\hat{\bm{R}}_{o}, \hat{\bm{t}}_{o})=\mathop{\textrm{argmax}}\limits_{\left\{(\hat{\bm{R}}_{o}^{(k)}, \hat{\bm{t}}_{o}^{(k)})|k=1,\cdots,K\right\}}~~s_{o}^{(k)}~.
\end{equation}

%In the rest of this section, we will describe the details about the proposed PRN and MMF-GCN modules.

\begin{figure}[!t]
	\centering
	\includegraphics[width=1.0\linewidth]{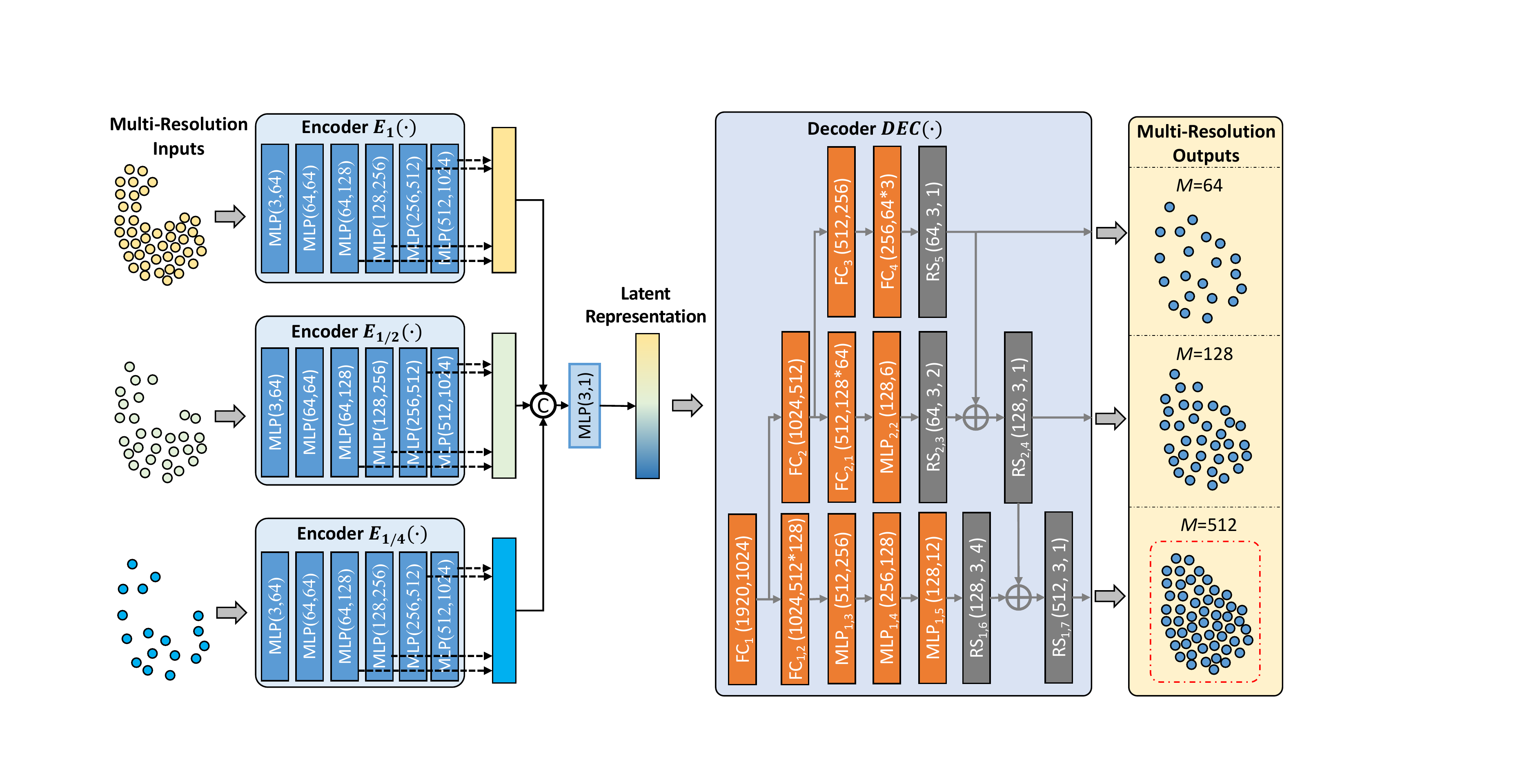}
	\caption{Detailed structure of PRN ($\bigoplus$: additive operation).}
	\label{fig:prn}
%\vspace{-0.4cm}
\end{figure}

\subsection{Point Refinement Network}

Recall that PRN aims to generate the refined 3D point cloud $\hat{\bm{P}}_{o}$ from the raw one of a low quality $\bm{P}_{o}$. As in Fig.~\ref{fig:prn}, PRN is endowed with an encoder-decoder architecture. To deal with the change in point density (resolution), we downsample $\bm{P}_{o}$ with $1/2$ and $1/4$ scales, resulting in two extra point clouds: $\bm{P}_{o,1/2}\in \mathbb{R}^{\frac{N}{2}\times 3}$ and $\bm{P}_{o,1/4}\in \mathbb{R}^{\frac{N}{4}\times 3}$. 

Accordingly, the encoder $E(\cdot)$ has three branches $E_{1}(\cdot)$, $E_{1/2}(\cdot)$, $E_{1/4}(\cdot)$, whose inputs are $\bm{P}_{o}$, $\bm{P}_{o,1/2}$, $\bm{P}_{o,1/4}$ and outputs are the three representations $\bm{v}_{1}=E_{1}(\bm{P}_{o})$, $\bm{v}_{1/2}=E_{1/2}(\bm{P}_{o,1/2})$, $\bm{v}_{1/4}=E_{1/4}(\bm{P}_{o,1/4})$. Each branch is a stack of six MLP layers. The concatenation of $\bm{v}_{1}$, $\bm{v}_{1/2}$ and $\bm{v}_{1/4}$ followed by an MLP layer forms the intermediate latent representation $\bm{v}$. 

%The decoder $DEC(\cdot)$ employs a multi-resolution structure as in \cite{pcn}. The first branch with four fully-connected (FC) layers is built to obtain a low-resolution point cloud $\hat{\bm{P}}_{o,1/8}=FC_{4}(FC_{3}(FC_{2}(FC_{1}(\bm{v}))))\in \mathbb{R}^{\frac{M}{8}\times 3}$. The second branch with the first two shared FC layers and the two additional FC layers generates a mediate-resolution point cloud: $\hat{\bm{P}}_{o,1/4}=FC_{2,2}(FC_{2,1}(FC_{2}(FC_{1}(\bm{v}))))\in \mathbb{R}^{\frac{M}{4}\times 3}$. Then, $\hat{\bm{P}}_{o,1/8}$ is integrated into $\hat{\bm{P}}_{o,1/4}$ by an additive operation: $\hat{\bm{P}}_{o,1/4}:=\hat{\bm{P}}_{o,1/4}+\hat{\bm{P}}_{o,1/8}$. Similarly, the third branch finally renders the high-resolution point cloud $\hat{\bm{P}}_{o}\in \mathbb{R}^{M\times 3}$ through sharing the first FC layer, adding three new FC layers and incorporating the mediate-resolution information
%  as $\hat{\bm{P}}_{o}=FC_{1,4}(FC_{1,3}(FC_{1,2}(FC_{1}(v))))+\hat{\bm{P}}_{o,1/4}$.

%\red{
The decoder $DEC(\cdot)$ employs a multi-resolution structure as \cite{pcn} does. The first branch with four fully-connected (FC) layers as well as one reshape (RS) operation is built to obtain the coarse point cloud of a low-resolution $\hat{\bm{P}}_{o,1/8}=RS_{5}(FC_{4}(FC_{3}(FC_{2}(FC_{1}(\bm{v})))))\in \mathbb{R}^{\frac{M}{8}\times 3}$ (or $\mathbb{R}^{\frac{M}{8}\times 3\times 1}$ equivalently). The second branch consists of the first two shared FC layers, one additional FC layer, one MLP layer and one RS operation, generating a mediate-resolution point cloud $\hat{\bm{P}}_{o,1/4}=RS_{2,3}(MLP_{2,2}(FC_{2,1}(FC_{2}(FC_{1}(\bm{v}))))).$ Then, $\hat{\bm{P}}_{o,1/8}$ is integrated into $\hat{\bm{P}}_{o,1/4}$ by a broadcasting additive operation $\hat{\bm{P}}_{o,1/4}:=\hat{\bm{P}}_{o,1/4}\bigoplus\hat{\bm{P}}_{o,1/8}$, and $\hat{\bm{P}}_{o,1/4}$ is reshaped into  $\hat{\bm{P}}_{o,1/4}:=RS_{2,4}(\hat{\bm{P}}_{o,1/4})\in \mathbb{R}^{\frac{M}{4}\times 3}$. Similarly, the third branch finally renders a high-resolution point cloud $\hat{\bm{P}}_{o}\in \mathbb{R}^{M\times 3}$ by sharing the first FC layer, adding one FC layer, three MLP layers and one RS operation, and incorporating the mediate-resolution information as $\hat{\bm{P}}_{o} =RS_{1,7}(RS_{1,6}(MLP_{1,5}(MLP_{1,4}(MLP_{1,3}
(FC_{1,2}(FC_{1}(\bm{v})\\)))))\bigoplus\hat{\bm{P}}_{o,1/4}).
$
%}

For training PRN, we develop a multi-resolution regression loss formulated as follows:
\begin{equation}\label{Eq.1.2}
\begin{array}{rl}
\mathcal{L}_{\rm{mr}}=d_{\rm C}(\hat{\bm{P}}_{o},\bm{P}_{o,GT})+\sum \limits_{r\in\{\frac{1}{8},\frac{1}{4}\}} d_{\rm C}(\hat{\bm{P}}_{o,r},\bm{P}_{o,GT})
\end{array}
\end{equation}
where $\bm{P}_{o,GT}=[\bm{p}_{o,GT}^{(1)};\cdots;\bm{p}_{o,GT}^{(H)}]\in \mathbb{R}^{H\times 3}$ is ground-truth point cloud of object $o$. $d_{\rm C}(\cdot,\cdot)$ is the Chamfer distance defined as $d_{\rm C}(\bm{P},\bm{Q}) = \frac{1}{M} \sum\limits_{i} \underset{j}{\rm min} \Vert \bm{p}^{(i)}-\bm{q}^{(j)} \Vert_2^2 + \frac{1}{N} \sum\limits_{j} \underset{i}{\rm min} \Vert \bm{q}^{(j)}-\bm{p}^{(i)} \Vert_2^2$, given $\bm{P}=[\bm{p}^{(m)}]_{m=1,\cdots,M}\in\mathbb{R}^{M\times 3}$ and $\bm{Q}=[\bm{q}^{(n)}]_{n=1,\cdots,N}\in\mathbb{R}^{N\times 3}$.

In Eq. \eqref{Eq.1.2}, $\hat{\bm{P}}_{o}$ is forced to fit the ground-truth in low-to-high resolutions when minimizing $\mathcal{L}_{\rm{mr}}$. In other words, $DEC(\cdot)$ is forced to predict high-quality points in multiple resolutions by a unified structure, and thus optimized with more supervision than the single-resolution case. Moreover, as shown in Fig.~\ref{fig:prn}, the high-resolution output $\hat{\bm{P}}_{o}$ integrates multi-resolution information from $\hat{\bm{P}}_{o,1/4}$ and $\hat{\bm{P}}_{o,1/8}$. As a consequence, PRN mitigates the incompleteness and decreases the noise of the raw 3D points.

%Despite the aforementioned advantages of $L_{mr}$, the multi-resolution loss fails to explore the point distributions of the ground-truth $P_{o,GT}$. To address this issue, we additionally employ the following adversarial loss:

Despite the aforementioned advantages of $\mathcal{L}_{\rm{mr}}$, it fails to perceive the global point distribution of $\bm{P}_{o,GT}$. We handle this problem by introducing the adversarial loss:
\begin{equation}\label{eq:adv_loss}
\small
\mathcal{L}_{\rm adv} = \sum\limits_{h=1}^{H}\log(D(\bm{p}_{o,GT}^{(h)})) + \sum\limits_{m=1}^{M}\log(1-D(\hat{\bm{p}}_{o}^{(m)})),
\end{equation}
where $D(\cdot)$ is the discriminator to classify whether a point belongs to $\bm{P}_{o,GT}$ (``real") or not (``fake"). By minimizing $\mathcal{L}_{\rm{adv}}$, PRN is expected to generate $\hat{\bm{P}}_{o}$ that captures holistic point distribution of $\bm{P}_{o,GT}$, benefiting the quality of $\bm{P}_{o}$.

The regularized multi-resolution regression loss is thus formulated as:
\begin{equation}\label{eq:prn_loss}
\small
\mathcal{L}_{\rm prn} =\sum \limits_{o} \left(\lambda\cdot \mathcal{L}_{\rm{adv}}+\beta\cdot \mathcal{L}_{\rm{mr}}\right),
\end{equation}
where $\lambda$ and $\beta$ are the trade-off hyper-parameters.

%\textbf{Reconstruction Loss.} The current method\cite{pcn, topnet} directly outputs the single scale complete point cloud, which is not accurate enough for the pose estimation. Therefore, we propose a multi-scale output to supervise the training of network, as shown in the following:

%\textbf{Adversarial Loss.} We adopt the efficient objective function as our adversarial loss, and use BCEloss as the adversarial loss, which is chosen by classify the right generated point cloud and fault generation result. Our PRNet defines a map: $X \rightarrow Y$, where X is the input point cloud, Y is the output point cloud. In detail, the losses for the generator and discriminator are:

%{\small
%	\begin{align}
%	\begin{split}
%		\label{Eq.2}
%		L_{\rm GAN} = \sum\limits_{i \in S}log(D(y_{i})) + %\sum\limits_{j \in S}log(1-D(F(x_{i})))
%    \end{split}
%    \end{align}}
%
%where $x_{i}\in X$, $y_{i}\in Y$, S is the dataset of $X, Y$. 

%\begin{figure*}[htbp]
%	\centering
%	\includegraphics[width=1.0\linewidth]{fig4-mmfgcn.pdf}
%	\caption{Point Refinement Module and MMFGCN module. (a) shows the network of Point Refinement, the input is sampled points, the output is the modified points; (b) shows the fusion process for the generated points and image features.}
%	\label{fig:mmfgcn}
%\end{figure*}

\subsection{Multi-Modal Fusion GCN} 
\label{section:MMF}

\begin{table*}[!t]
   \renewcommand\arraystretch{1.2}
   \centering
      \caption{Comparison with the state-of-the-arts in terms of ADD(-S) (\%) on Linemod. Symmetric objects are marked in bold. $*$/$\dag$ indicates that the method only uses real/synthetic data for training. PF and DF refer to PointFusion \cite{pointfusion} and DenseFusion \cite{densefusion}, respectively.}
   \label{tab:linemod_result}
   \resizebox{0.99\textwidth}{!}{%
%   \begin{tabular}{p{1.6cm}<{\raggedright}|p{1.4cm}<{\centering}|p{1.2cm}<{\centering}|p{1.0cm}<{\centering}|p{1.0cm}<{\centering}|p{1.0cm}<{\centering}|p{1.2cm}<{\centering}|p{1.2cm}<{\centering}|p{1.2cm}<{\centering}|p{1.0cm}<{\centering}|p{1.0cm}<{\centering}|p{1.0cm}<{\centering}|p{1.0cm}<{\centering}}
   \begin{tabular}{l|c|c|c|c|c|c|c|c|c|c|c|c}
      \hline
               & \multicolumn{5}{c|}{RGB based methods}                       & \multicolumn{7}{c}{RGB-D based methods}\\ \hline
%              Object & PoseCNN*\ \cite{POSECNN, deepim}& PVNet \ \cite{pvnet} & CDPN \ \cite{cpdn} & DPOD \ \cite{dpod} & DPVL \ \cite{DPVL}  & PF{*} \ \cite{pointfusion} & SSD6D$^{\dag}$ \cite{SSD-6D} & DF{*} \ \cite{densefusion} & PVN3D \cite{pvn3d} & G2L{*} \ \cite{G2L} &Ours{*}  & Ours \\ 
Object & PoseCNN*\ & PVNet & CDPN & DPOD & DPVL & PF{*} & SSD6D$^{\dag}$ & DF{*} & PVN3D & G2L{*} & Ours{*}  & Ours \\ 
                & \cite{POSECNN, deepim}& \cite{pvnet} &  \cite{cpdn} & \cite{dpod} & \cite{DPVL}  &  \cite{pointfusion} &  \cite{SSD-6D} & \cite{densefusion} &  \cite{pvn3d} &  \cite{G2L} &   &  \\ \hline
   ape         & 77.0           & 43.6  & 64.4 & 87.7  & 69.1  & 70.4                                                         & 65.0                                                       & 92.3                                                         & 97.3  & 96.8 & 97.6 & \textbf{99.2}  \\ 
   benchvise   & 97.5           & 99.9  & 97.8 & 98.5  & \textbf{100.0} & 80.7                                                         & 80.0                                                       & 93.2                                                         & 99.7  & 96.1 & 99.2 & 99.8  \\ 
   camera      & 93.5           & 86.9  & 91.7 & 96.1  & 94.1  & 60.8                                                         & 78.0                                                       & 94.4                                                         & 99.6  & 98.2 & 99.4 & \textbf{100.0}  \\ 
   can         & 96.5           & 95.5  & 95.9 & \textbf{99.7}  & 98.5  & 61.1                                                         & 86.0                                                       & 93.1                                                         & 99.5  & 98.0 & 98.4 & 99.4  \\ 
   cat         & 82.1           & 79.3  & 83.8 & 94.7  & 83.1  & 79.1                                                         & 70.0                                                       & 96.5                                                         & \textbf{99.8}  & 99.2 & 98.7 & \textbf{99.8} \\ 
   driller     & 95.0           & 96.4  & 96.2 & 98.8  & 99.0  & 47.3                                                         & 73.0                                                       & 87.0                                                         & 99.3  & \textbf{99.8}  & 98.8 & \textbf{99.8}  \\ 
   duck        & 77.7           & 52.6  & 66.8 & 86.3  & 63.5  & 63.0                                                         & 66.0                                                       & 92.3                                                         & 98.2  & 97.7 & \textbf{98.9} & 98.7  \\ 
   \textbf{eggbox}      & 97.1           & 99.2  & 99.7 & 99.9  & \textbf{100.0} & 99.9                                                         & \textbf{100.0}                                                      & 99.8                                                         & 99.8  & \textbf{100.0} & 99.9 & 99.6  \\ 
   \textbf{glue}        & 99.4           & 95.7  & 99.6 & 96.8  & 98.0  & 99.3                                                         & \textbf{100.0}                                                      & \textbf{100.0}                                                        & \textbf{100.0} & \textbf{100.0} & \textbf{100.0} &  \textbf{100.0}  \\
   holepuncher & 52.8           & 82.0  & 85.8 & 86.9  & 88.2  & 71.8                                                         & 49.0                                                       & 92.1                                                         & \textbf{99.9}  & 99.0 & 99.4 & 99.8  \\ 
   iron        & 98.3           & 98.9  & 97.9 &\textbf{ 100.0} & 99.9  & 83.2                                                         & 78.0                                                       & 97.0                                                         & 99.7  & 99.3 & 98.5 & 99.5  \\ 
   lamp        & 97.5           & 99.3  & 97.9 & 96.8  & 99.8  & 62.3                                                         & 73.0                                                       & 95.3                                                         & 99.8  & 99.5 & 99.2 & \textbf{100.0}  \\ 
   phone       & 87.7           & 92.4  & 90.8 & 94.7  & 96.4  & 78.8                                                         & 79.0                                                       & 92.8                                                         & 99.5  & 98.9 & 98.4 & \textbf{99.7}  \\ \hline
   MEAN         & 88.6           & 86.3  & 89.9 & 95.2  & 91.5  & 73.7                                                         & 79.0                                                       & 94.3                                                         & 99.4  & 98.7  & 98.9 & \textbf{99.6}  \\ \hline
   \end{tabular}
   }
   %\vspace{-0.4cm}
   \end{table*}

%\begin{figure}[!t]
%	\centering
%	\includegraphics[width=1.0\linewidth]{fuse.pdf}
%	\caption{The detailed description of graph fuse module and node feature update process. (a) The node feature update by using function $fuse_{j}^{*} = \sum{(fuse_{j, i} - fuse_{j}, fuse_{j})}$, where $i \in K$, K representes the K-NN (K-NearestNeighbor) node, selected by Euclidean distance in feature space; (b) The node of first Graph, coordinates (X, Y, Z) is used directly; (c) Embedding geometry feature is used for the rest graph, M is the dimension of geometry feature, N is the dimension of texture feature.}
%	\label{fig:fuse-detail}
%\end{figure}

As mentioned before, given the RGB ($I_{o,rgb}$) and point cloud ($\bm{P}_{o}$ and $\hat{\bm{P}}_{o}$) data of object $o$, MMF-GCN integrates multi-modal information into more effective representation ($\bm{G}_{o}$) for accurate 6D pose estimation. 

Specifically, MMF-GCN first extracts the geometry feature $\bm{f}_{o,d}^{(i)}$ from $\bm{P}_{o}$ and the texture feature $\bm{f}_{o,rgb}^{(i)}$ from $I_{o, rgb}$ for the $i$-th point $\bm{p}^{(i)}_{o}\in \bm{P}_{o}$. The normalized coordinate of $\bm{p}_{o}^{(i)}$ is directly used as the geometry feature, and by mapping this coordinate to the pixel on $I_{o, rgb}$, PSPNet \cite{pspnet} with the ResNet-18 backbone is adopted to compute pixel-wise representation as the texture feature.

When $\{\bm{f}_{o,rgb}^{(i)}\}$ and $\{\bm{f}_{o,d}^{(i)}\}$ are ready, a $k$-Nearest Neighbor ($k-$NN) graph $\mathcal{G}_{f}=(\mathcal{V}_{f}, \mathcal{E}_{f})$ is constructed. $\mathcal{V}_{f}=\{\bm{p}_{o}^{(1)},\cdots,\bm{p}_{o}^{(N)}\}$ and $\mathcal{E}_{f}=\{(\bm{p}_{o}^{(i)},\bm{p}_{o}^{(j)})|\bm{p}_{o}^{(j)}\in \mathcal{N}_{k}(\bm{p}_{o}^{(i)})\}$ denote the vertices and the edges, and $\mathcal{N}_{k}(\bm{p}_{o}^{(i)})$ indicates the $k$ nearest neighbors of $\bm{p}_{o}^{(i)}$. The edge feature is defined as $\bm{e}^{(i,j)}=h_{\bm{\theta}}(\bm{f}_{o}^{(i)}-\bm{f}_{o}^{(j)}, \bm{f}_{o}^{(i)})$ with $\bm{f}_{o}^{(i)}=[\bm{f}_{o,rgb}^{(i)},\bm{f}_{o,d}^{(i)}]$, where $h_{\bm{\theta}}(\cdot,\cdot)$ is a nonlinear function parameterized by $\bm{\theta}$.

Afterwards, a graph convolution network $GCN_{f}(\cdot)$ is employed to capture local inter-modality correlations, with EdgeConv \cite{DGCNN} for graph convolutions. The basic updating scheme is formulated as:
\begin{equation*}
\bm{g}_{o,f}^{(i,l)}=MP\left(h_{\bm{\theta}^{(l-1)}}\left(\bm{g}_{o,f}^{(i,l-1)}-\bm{g}_{o,f}^{(j,l-1)}, \bm{g}_{o,f}^{(i,l-1)}\right)\right),
\end{equation*}
where $\bm{g}_{o,f}^{(i,l)}$ denotes the $i$-th edge feature in the $l$-th layer, $h_{\bm{\theta}^{(l-1)}}(\cdot,\cdot)$ is a nonlinear function in the $(l-1)$-th layer, and $MP(\cdot)$ refers to max pooling. The representation $\bm{G}_{o,f}=[\bm{g}_{o,f}^{(j)}]_{j=1,\cdots,J} \in \mathbb{R}^{J\times d_{f}}$ is then obtained, where $\bm{G}_{o,f}=GCN_{f}(\{\bm{f}_{o,rgb}^{(i)}, \bm{f}_{o,d}^{(i)}\})$; $J$ and $d_{f}$ are the point number and the feature dimension, respectively. 

As $\bm{P}_{o}$ is usually incomplete and noisy, MMF-GCN encodes the geometry attribute of $\hat{\bm{P}}_{o}$ and incorporates it into $\bm{G}_{o,f}$ as a complement.
% Since the raw 3D points $\bm{P}_{o}$ is usually incomplete and noisy, MMF-GCN encodes the geometry attributes of its refinement $\hat{\bm{P}}_{o}$, and incorporate it into $\bm{G}_{o,f}$ as a complement.
Concretely, similar to $\mathcal{G}_{f}$, another $k$-NN graph $\mathcal{G}_{ref}$ is built based on $\hat{\bm{P}}_{o}$, and an extra GCN $GCN_{ref}(\cdot)$ is employed. The refined geometry feature is calculated using EdgeConv:  $\bm{G}_{o,ref}=[\bm{g}_{o,ref}^{(j)}]_{k=1,\cdots,J}\in \mathbb{R}^{J\times d_{ref}}$, where $\bm{G}_{o,ref}=GCN_{ref}(\hat{\bm{P}}_{o})$ and $d_{ref}$ is the feature dimension. $\bm{G}_{o,ref}$ is subsequently combined with $\bm{G}_{o,f}$ through simple concatenation, which is further integrated by a few stacked FC layers $T(\cdot)$. At last, multi-modal representation is formed as $\bm{G}_{o}=T([\bm{G}_{o,r}, \bm{G}_{o,ref}])$ for 6D pose estimation.

\subsection{Training Objectives}

The objective function for training PR-GCN consists of two parts: the pose estimation loss $\mathcal{L}_{\rm{pose}}$ and the regularized multi-resolution regression loss $\mathcal{L}_{\rm{prn}}$ as depicted in Eq. \eqref{eq:prn_loss}. 

Given the ground-truth 6D pose $(\bm{R}_{o},\bm{t}_{o})$ and the predictions $\{(\hat{\bm{R}}_{o}^{(k)},\hat{\bm{t}}_{o}^{(k)}, s^{(k)}_{o})\}$ at $K$ points $\{\bm{x}_{o}^{(k)}\}$, the pose estimation error of the $i$-th prediction $(\hat{\bm{R}}_{o}^{(i)},\hat{\bm{t}}_{o}^{(i)})$ is defined as $
e_{o}^{(i)} = \frac{1}{K}\sum_{j=1}^{K}\min_{k}\|(\bm{R}_{o}\bm{x}_{o}^{(j)}+\bm{t}_{o})-(\hat{\bm{R}}_{o}^{(i)} \bm{x}^{(k)}_{o} + \hat{\bm{t}}_{o}^{(i)})\|^2_{2}$. Based on $e_{o}^{(i)}$, we adopt an extra regularization term on the prediction scores $\{s^{(i)}_{o}\}$ as in \cite{densefusion} and formulate the pose estimation loss as: 
\begin{equation}
\label{pose_all}
\mathcal{L}_{\rm{pose}} = \frac{1}{K}\sum_{o}\sum_{i}e^{(i)}_{o}\cdot \left(s^{(i)}_{o}-\log(s^{(i)}_{o})\right).
\end{equation}

By combining Eq. \eqref{pose_all} and Eq. \eqref{eq:prn_loss}, the overall training objective function is written as:

\begin{equation}
\label{loss}
\mathcal{L} = \mathcal{L}_{\rm{pose}} + \mu\cdot \mathcal{L}_{\rm{prn}},
\end{equation}
where $\mu$ is the trade-off hyper-parameter.
%-------------------------------------------------------------------------

%-------------------------------------------------------------------------
\section{Experiments}

\subsection{Datasets and Metrics}
Extensive evaluation is made on three datasets: Linemod \cite{linemod}, Occlusion Linemod \cite{brachmann2014learning} and YCB-Video \cite{POSECNN}. 

%------------------------------------------------------------------------

%\subsection{Datasets and Evaluatin Metrics}

\textbf{Linemod} \cite{linemod} is composed of 15 RGB-D videos of 15 low-textured objects. Following \cite{bb8}, 13 objects are considered and the standard training/testing split is adopted as in \cite{densefusion, POSECNN}. \textbf{Occlusion Linemod} is collected by annotating a subset of Linemod (8 out of 15 objects), where each image has multiple occluded objects, making it more challenging. \textbf{YCB-Video} \cite{POSECNN} includes 21 objects with various textures and sizes. It provides RGB-D images and detailed pose annotations. There are 130K real images from 92 videos and 80K synthetically rendered ones, and 16,189 real and all the synthesized images are used in training, according to \cite{densefusion}. 
%Since \cite{pvn3d, pvnet} generate synthesis training images for boosting the performance, we additionally report the result by training our method with synthesis images. 

%Their definitions are as follows:

%\begin{equation}
%\label{eqn:ADD}
%\textrm{ADD} = \frac{1}{m} \sum_{x \in o} || (Rx+t) - (R^*x + t^*) ||
%\end{equation}

%where $m$ denotes the number of the vertexes on the object, $x$ denotes the vertex, $R$ and $R*$ denotes ground-truth and predicted rotation matrix, $t$ and $t*$ denotes the ground-truth and predicted translation matrix. $\mathbf{O}$ denotes the testing set.

%------------------------------------------------------------------------

%Based on ADD metric \cite{linemod}, we can furtherly transform the model points by the estimated $[R^*|t^*]$ and the ground truth poses $[R|t]$, respectively. Then, we calculate the average distance of all mesh points.

%\begin{equation}
%\label{eqn:ADDS}
%\textrm{ADD{-}S} = \frac{1}{m} \sum_{x_1 \in o} \min_{x_2 \in o}{|| (Rx_1+t) - (R^*x_2 + t^*) ||}
%\end{equation}

%The ADD-S metric \cite{linemod} calculates the mean average distance from each mesh point to its closest point transformed by  $[R^*|t^*]$. 

\begin{table*}[!t]
\centering
\caption{Comparison of ADD(-S) AUC (\%) on Occlusion Linemod. Symmetric objects are marked in bold.}
\label{table:occ-lin}
 \resizebox{0.99\textwidth}{!}{
\begin{tabular}{l|c|c|c|c|c|c|c|c}
\hline
%\multirow{2}{*}{Object}     & PoseCNN & DeepHeat & SS & Pix2pose & PVNet  & HybridPose & PVN3D & Ours \\ 
% & \cite{POSECNN} &  \cite{DBLP:conf/eccv/OberwegerRL18} &  \cite{DBLP:conf/cvpr/HuF0S20} & \cite{pix2pose} & \cite{pvnet} & \cite{DBLP:conf/cvpr/SongSH20} & \cite{pvn3d} & \\
Object     & PoseCNN \cite{POSECNN} & DeepHeat  \cite{DBLP:conf/eccv/OberwegerRL18} & SS \cite{DBLP:conf/cvpr/HuF0S20} & Pix2pose \cite{pix2pose} & PVNet \cite{pvnet}  & HybridPose \cite{DBLP:conf/cvpr/SongSH20} & PVN3D \cite{pvn3d} & Ours \\ 
\hline
Ape         & 9.6    & 12.1      & 17.6      & 22.0     & 15.8  & 20.9       & 33.9  &  \textbf{40.2}    \\
Can         & 45.2   & 39.9      & 53.9      & 44.7     & 63.3  & 75.3       & \textbf{88.6}  &  76.2    \\
Cat         & 0.9    & 8.2       & 3.3       & 22.7     & 16.7  & 24.9       & 39.1  &  \textbf{57.0}    \\
Driller     & 41.4   & 45.2      & 62.4      & 44.7     & 65.7  & 70.2       & 78.4  &  \textbf{82.3}    \\ 
Duck        & 19.6   & 17.2      & 19.2      & 15.0     & 25.2  & 27.9       & \textbf{41.9}  &  30.0    \\ 
\textbf{Eggbox}      & 22.0   & 22.1      & 25.9      & 25.2     & 50.2  & 52.4       & \textbf{80.9}  &  68.2    \\
\textbf{Glue}        & 38.5   & 35.8      & 39.6      & 32.4     & 49.6  & 53.8       & \textbf{68.1}  & 67.0    \\ 
Holepuncher & 22.1   & 36.0      & 21.3      & 49.5     & 39.7  & 54.2       & 74.7  &  \textbf{97.2}    \\
\hline
MEAN        & 24.9   & 27.0      & 27.0      & 32.0     & 40.8  & 47.5       & 63.2  & \textbf{65.0} \\ 
\hline
\end{tabular}
}
%\vspace{-0.2cm}
\end{table*}

\begin{table*}[!thp]
   \renewcommand\arraystretch{1.2}
   \centering
   \caption{Comparison of AUC (\%) and ADD-S \textless{} 2cm (\%) (``\textless{}2cm'' for short) on YCB-Video. Symmetric objects are highlighted in bold.}
   \label{tab:ycb_result}
   \resizebox{0.99\textwidth}{!}{%
   \centering
%   \begin{tabular}{l|p{1.2cm}<{\centering}|p{0.8cm}<{\centering}|p{0.8cm}<{\centering}|p{0.8cm}<{\centering}|p{0.8cm}<{\centering}|p{0.8cm}<{\centering}|p{0.8cm}<{\centering}|p{0.8cm}<{\centering}|p{0.8cm}<{\centering}|p{0.8cm}<{\centering}|p{0.8cm}<{\centering}}
\begin{tabular}{l|c|c|c|c|c|c|c|c|c|c|c}
   \hline
    & \multicolumn{2}{c|}{PoseCNN+ICP \ \cite{POSECNN}} & \multicolumn{2}{c|}{DenseFusion \ \cite{densefusion}} & \multicolumn{2}{c|}{PVN3D \ \cite{pvn3d}} &  \multicolumn{2}{c|}{CF \ \cite{CF} } & \multicolumn{1}{c|}{G2L  \  \cite{G2L}} &\multicolumn{2}{c}{Ours} 
%    & \multicolumn{2}{c|}{ \cite{POSECNN}} & \multicolumn{2}{c|}{ \cite{densefusion}} & \multicolumn{2}{c|}{ \cite{pvn3d}} &  \multicolumn{2}{c|}{ \cite{CF}} & \multicolumn{1}{c|}{  \cite{G2L}} &\multicolumn{2}{c}{ }
    \\ \hline
    & AUC       & \textless{}2cm    & AUC       & \textless{}2cm    & AUC  & \textless{}2cm   & AUC  & \textless{}2cm  & AUC  & AUC    & \textless{}2cm \\ 
    \hline
   002\_master\_chef\_can   & 95.8 & \textbf{100.0} & 96.4 & \textbf{100.0}  & 96.0 & \textbf{100.0} & 92.5  & 98.7 & 94.0  & \textbf{97.1} & \textbf{100.0}  \\ 
   %\hline
   003\_cracker\_box        & 92.7 & 91.6  & 95.5 & 99.5  & 96.1 & \textbf{100.0} & 95.4  & 98.6 & 88.7  & \textbf{97.6} & \textbf{100.0} \\ 
   %\hline
   004\_sugar\_box          & 98.2 & \textbf{100.0}  & 97.5 & \textbf{100.0}  & 97.4 & \textbf{100.0} & 96.7  & 99.9 & 96.0 &\textbf{98.3} & \textbf{100.0} \\ 
   %\hline
   005\_tomato\_soup\_can   & 94.5 & 96.9  & 94.6 & 96.9  & \textbf{96.2} & \textbf{98.1} & 92.0  & 95.8 & 86.4  & 95.3 & 97.6   \\ 
   %\hline
   006\_mustard\_bottle     & \textbf{98.6} & \textbf{100.0}  & 97.2 & \textbf{100.0} & 97.5 & \textbf{100.0} & 94.8 & 97.5 & 95.9 & 97.9 & \textbf{100.0} \\ 
   %\hline
   007\_tuna\_fish\_can      & 97.1 & \textbf{100.0} & 96.6 & \textbf{100.0} & 96.0 & \textbf{100.0} & 88.8 & 84.1 & 84.1 & \textbf{97.6} & \textbf{100.0} \\ 
   %\hline
   008\_pudding\_box        & 97.9 & \textbf{100.0} & 96.5 & \textbf{100.0} & 97.1 & \textbf{100.0} & 93.2 & 98.6 & 93.5  & \textbf{98.4} & \textbf{100.0} \\ 
   %\hline
   009\_gelatin\_box        & \textbf{98.8} & \textbf{100.0}  & 98.1 & \textbf{100.0}  &97.7 & \textbf{100.0} & 95.7 & \textbf{100.0} & 96.8  &  96.2 & 94.4 \\ 
   %\hline
   010\_potted\_meat\_can   & 92.7 & 93.6  & 91.3 & 93.1  &93.3 & 94.6 & 86.2 & 83.9 & 86.2 & \textbf{96.6} & \textbf{99.1} \\ %\hline
   011\_banana              & 97.1 & 99.7  & 96.6 & \textbf{100.0} & 96.6 & \textbf{100.0} & 92.6 & 98.9 & 96.3 & \textbf{98.5} & \textbf{100.0} \\ 
   %\hline
   019\_pitcher\_base       & 97.8 & \textbf{100.0} & 97.1 & \textbf{100.0} & 97.4 & \textbf{100.0} & 95.4 & 98.4 & 91.8  & \textbf{98.1} & \textbf{100.0} \\ 
   %\hline
   021\_bleach\_cleanser    & 96.9 & 99.4  & 95.8 & \textbf{100.0} & 96.0 & \textbf{100.0} & 89.0 & 86.2 & 92.0  & \textbf{97.9} & \textbf{100.0} \\ 
   %\hline
   \textbf{024\_bowl}    & 81.0 & 54.9 & 88.2 & \textbf{98.8}  & 90.2 & 80.5 & 86.1 & 94.3 & 86.7  & \textbf{90.3} & 96.6\\ %\hline
   025\_mug                 & 95.0 & 99.8 & 97.1 & \textbf{100.0} & 97.6 & \textbf{100.0} & 93.5 & 94.8 & 95.4  & \textbf{98.1} & \textbf{100.0}\\ 
   %\hline
   035\_power\_drill        & \textbf{98.2} & 99.6 & 96.0 & 98.7 & 96.7 & \textbf{100.0} & 82.9 & 84.8 & 95.2  & 98.1 & \textbf{100.0}\\ 
   %\hline
   \textbf{036\_wood\_block}    & 87.6 & 80.2 & 89.7 & 94.6 & 90.4 & 93.8 & 92.3 & 99.6 & 86.2  & \textbf{96.0} & \textbf{100.0}\\ %\hline
   037\_scissors            & 91.7 & 95.6  & 95.2 & \textbf{100.0} & \textbf{96.7} & \textbf{100.0} & 90.1 & 89.5 & 83.8  & \textbf{96.7} & \textbf{100.0}\\ 
   %\hline
   040\_large\_marker       & 97.2 & 99.7 & 97.5 & \textbf{100.0} & 96.7 & 99.8 & 93.9 & 99.8 & 96.8  & \textbf{97.9} & \textbf{100.0} \\ 
   %\hline
   \textbf{051\_large\_clamp}        & 75.2 & 74.9  & 72.9 & 79.2  & 93.6 & \textbf{93.6}  & 70.3  & 76.7 & \textbf{94.4}  & 87.5 & 93.3  \\ 
   %\hline
   \textbf{052\_extra\_large\_clamp} & 64.4 & 48.8  & 69.8 & 76.3  & 88.4 & 83.6 & 69.5 & 74.5 & \textbf{92.3}  & 79.7 & \textbf{84.6}  \\ 
   %\hline
   \textbf{061\_foam\_brick}         & 97.2 & \textbf{100.0}  & 92.5 & \textbf{100.0} & 96.8 & \textbf{100.0} & 94.6  & \textbf{100.0} & 94.7  & \textbf{97.8} & \textbf{100.0}\\ 
   \hline
   MEAN     & 93.0 & 93.2 & 93.1 & 96.8  & 95.5 & 97.6 & 89.8  & 93.1 & 92.4  & \textbf{95.8} & \textbf{98.5}\\ \hline
   \end{tabular}%
   }
   %\vspace{-0.2cm}
\end{table*}

As in the literature, two main metrics are employed for evaluation, \ie, Average Distance (ADD) \cite{POSECNN} and ADD-Symmetric (ADD-S) \cite{POSECNN}, designed for general objects and symmetric objects, respectively. DenseFusion \cite{densefusion} gives the ADD-S smaller than 2 centimeters (ADD-S$<$2cm) for real applications \eg robotic manipulation, and PoseCNN \cite{POSECNN} and DenseFusion \cite{densefusion} report the Area Under the ADD-S Curve (AUC) with the maximum threshold at 0.1m. We also show them for comparison.

\subsection{Implementation Details}
\label{section4.1}
We fix the size of RGB images as 480$\times$640. The numbers of raw/refined 3D points, \ie, $N$/$M$, are set to 100/512 and 100/1,024 on Linemod (Occlusion Linemod) and YCB-Video, respectively. In MMF-GCN, refined point clouds are down-sampled to 100 points by FPS. When building the graphs $\mathcal{G}_{f}$ and $\mathcal{G}_{ref}$, we utilize 30 nearest neighbors, \ie, $k=30$. In PRN, the hyper-parameters $\lambda$, $\beta$ and $\mu$ in the overall training loss $\mathcal{L}$ are set to 0.05, 0.95 and 1.0. To train PR-GCN in a more stable way, PRN and MMF-GCN are progressively optimized. For instance, on YCB-Video, PRN and MMF-GCN are first alternatively trained for 15 epochs and then jointly optimized for 30 epochs.

In PRN training, we adopt the ADAM optimizer with the learning rate of 0.0001 and the batch size of 48. The rest parts of PR-GCN are trained for 40, 20 and 60 epochs on Linemod, Occlusion Linemod and YCB-Video, respectively, where the learning rate is initially set to 0.0001 and decayed by a factor of 0.3 after half of the maximal epochs. %All experiments are implemented on {\color{red}an} Nvidia GTX 1080Ti GPU.

\subsection{Comparison with the State-of-the-art Methods}

\textbf{Results on Linemod.} We first compare PR-GCN to the state-of-the-art methods on Linemod, including the RGB based models: PoseCNN (+DeepIM) \cite{POSECNN, deepim}, PVNet \cite{pvnet}, CDPN \cite{cpdn}, DPOD \cite{dpod} and DPVL \cite{DPVL} and the RGB-D based ones: Point Fusion \cite{pointfusion}, SSD6D (+ICP) \cite{SSD-6D}, Dense Fusion \cite{densefusion}, PVN3D \cite{pvn3d} and G2L\cite{G2L}. Several approaches, denoted by `$*$' or `$\dag$' in Table \ref{tab:linemod_result}, only adopt real or synthetic training data, whist the others use both. In our work, we mainly consider the setting with the two types of training data, and also report the performance with real data only.

Table \ref{tab:linemod_result} summarizes the ADD(-S) of different methods on Linemod, and we can see that the RGB-D based deep models (\emph{e.g.}, PVN3D and G2L) remarkably outperform the RGB based ones by a large margin due to additional geometry information given by the depth channel. Regarding the RGB-D counterparts, the proposed PR-GCN achieves better performance, which improves PointFusion and DenseFusion by 25.2\% and 4.6\%, respectively. Our approach also boosts the performance of keypoint-based methods, including PVN3D and G2L. It is worth noting that the second best method, \emph{i.e.} PVN3D, uses 70,000 synthetic training data, and needs to train different models for distinct object categories. In contrast, our method trains a universal model for all object categories, and merely utilizes 3,500 synthetic training data, which is much more efficient than PVN3D.

\begin{figure}[!t]
  \centering
  \subfigure[]{
    \label{fig:subfig:a} %% label for first subfigure
    \includegraphics[width=\linewidth]{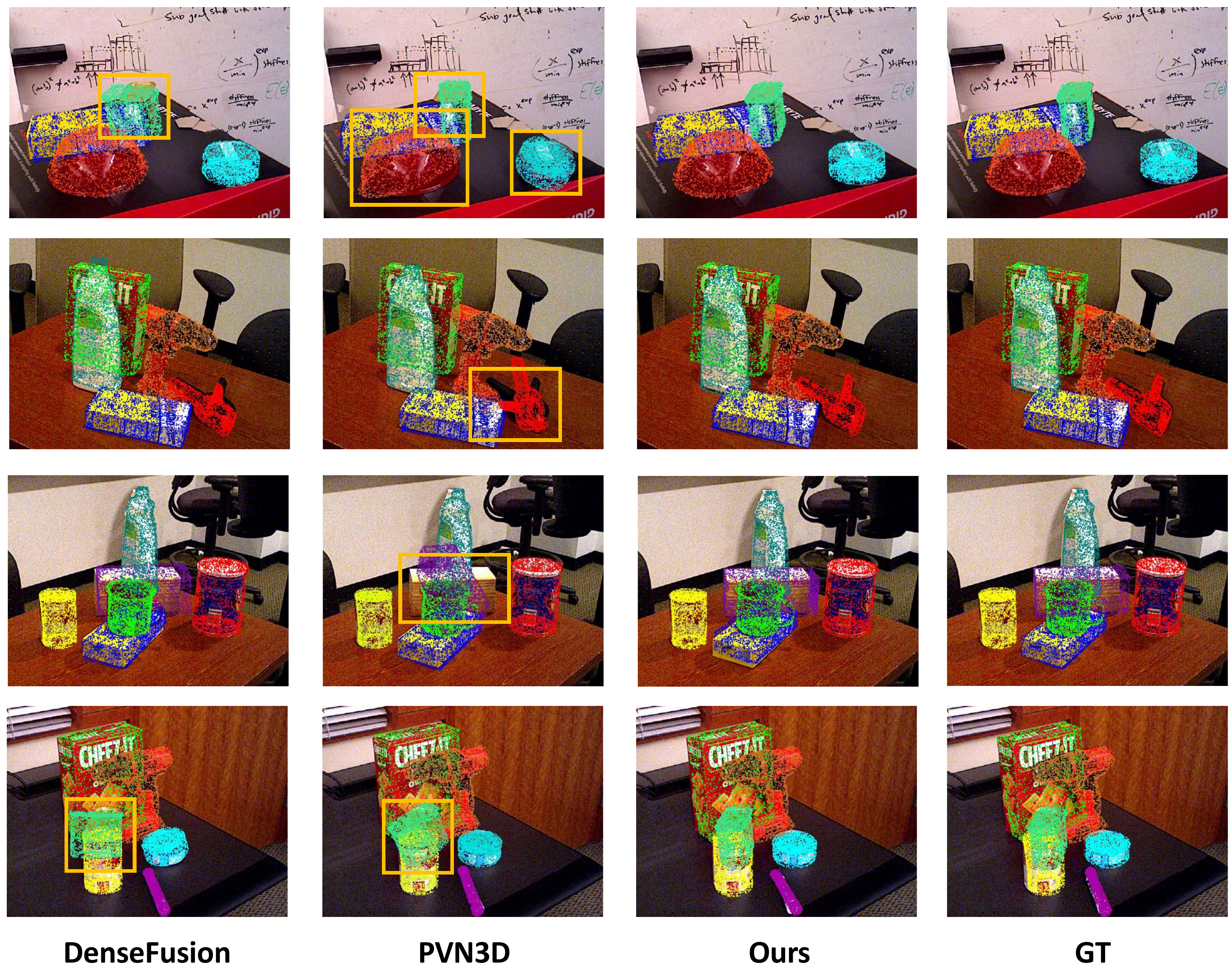}}
  %\hspace{1in}
  \subfigure[]{
    \label{fig:subfig:b} %% label for second subfigure
    \includegraphics[width=\linewidth, height=0.33\linewidth]{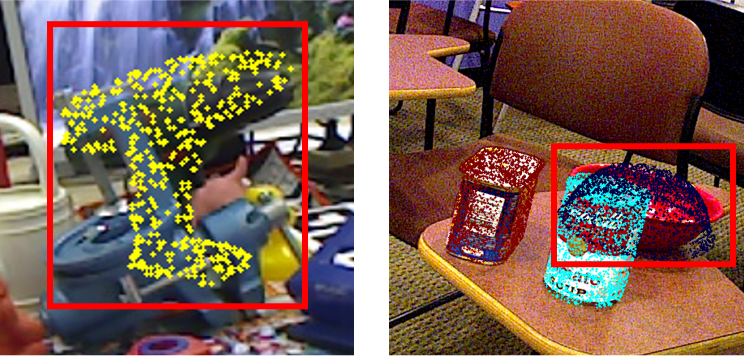}}
  \caption{\textbf{Qualitative analysis}. (a) Visualization results on YCB-Video. From left to right: provided by DenseFusion (with 2 iterations), PVN3D, PR-GCN (ours) and ground-truth (GT). Orange bounding boxes highlight inaccurate estimation. (b) Failure cases. Left: heavy occlusion (`Holepuncher' from Occlusion Linemod) and right: symmetry object (`Bowl' from YCB-Video).}
  \label{fig:vis_res} %% label for entire figure
     %\vspace{-0.2cm}
\end{figure}

%Since some existing works don't use extra synthetic data during training, we also report ADD-S by PR-GCN without synthetic data, which is still much higher than compared methods.

\textbf{Results on Occlusion Linemod.}
To evaluate the robustness of PR-GCN to inter-object occlusions, we display detailed results on Occlusion Linemod, in comparison with PoseCNN \cite{POSECNN}, DeepHeat \cite{DBLP:conf/eccv/OberwegerRL18}, SS \cite{DBLP:conf/cvpr/HuF0S20}, Pix2Pose \cite{pix2pose}, PVNet \cite{pvnet}, HybridPose \cite{DBLP:conf/cvpr/SongSH20} and PVN3D \cite{pvn3d}. As shown in Table \ref{table:occ-lin}, our method consistently reaches the top ADD(-S) AUC and achieves the best mean ADD(-S) AUC, highlighting its superiority in the presence of heavy occlusions.

%\begin{figure}[!t]
%	\centering
%	\includegraphics[width=\linewidth]{disp_cmp_6d_pose.pdf}
%	\caption{\textbf{Visualization of the pose estimation results on YCB-Video.} From left to right: the results by using DenseFusion (with 2 iterations), PVN3D, our method and the ground truth (GT). Orange bounding boxes highlight inaccurate estimations.}
%	\label{fig:vis-ycb}
%	\vspace{-0.4cm}
%\end{figure}

\textbf{Results on YCB-Video.} We then extend our analysis on this database and compare PR-GCN with PoseCNN (+ICP) \cite{POSECNN}, DenseFusion \cite{densefusion}, PVN3D \cite{pvn3d}, CF \cite{CF} and G2L \cite{G2L}. Table \ref{tab:ycb_result} shows the AUC and ADD-S\textless{}2cm for various methods. It can be observed that our method achieves the highest performance on both the metrics. For instance, compared to PVN3D and DenseFusion, PR-GCN improves the ADD-S$<$2cm by 0.9\% and 1.7\%, respectively. 
%To make a fair comparison with PVN3D, we use the same segmentation result.
%All the methods take the same bounding box and segmentation mask as the input for fair comparison.

\textbf{Qualitative results.} We additionally provide qualitative results in Fig.~\ref{fig:subfig:a}, comparing to DenseFusion and PVN3D. Due to cluttered backgrounds and severe occlusions, DenseFusion and PVN3D predict inaccurate poses in many cases, while our PR-GCN performs more robustly with much better results. We also demonstrate failure cases in Fig.~\ref{fig:subfig:b}, revealing that PR-GCN fails when dealing with extremely occluded objects and some symmetric ones. 

\textbf{Inference efficiency.} Besides the accuracy, we evaluate the efficiency of our method on Linemod. As shown in Table \ref{table:runtime}, each key component infers fast, and the full pipeline takes 68ms on an Nvidia 1080Ti GPU, which is acceptable in downstream tasks such as robotic grasping.

\begin{table}[!t]%\footnotesize
%\vspace{-0.2cm}
\centering
\caption{Inference time of Segmentation (Seg), Point Refinement (PR), Pose Estimation (PE) and full PR-GCN (Full) on Linemod.}
\resizebox{0.37\textwidth}{!}{%
\begin{tabular}{l|c|c|c|c}
\hline
%\multirow{1}{*}{} & \multicolumn{4}{c}{Ours}                 \\ \cline{2-5} 
Component                        & Seg & PR & PE & Full \\ \hline
\multicolumn{1}{c|}{Time (s)} &    0.030          &       0.008  & 0.030 & 0.068      \\ 
\hline
\end{tabular}
\vspace{0.2cm}
}
\label{table:runtime}
\end{table}

\begin{table}[!t]
	\centering
	\caption{Ablation study of PR-GCN in ADD(-S) (\%) on Linemod.}
	\resizebox{0.45\textwidth}{!}{%
   \label{tab:ablation1}
	\begin{tabular}{l|c|c|c}
		\hline
		Method & PRN  & MMF-GCN & MEAN  \\
		\hline
		Baseline (with DGCNN) & $\times$       & $\times$                & 94.8      \\
		Baseline+PRN & $\surd$       & $\times$                 & 96.8       \\
		Baseline+MMF-GCN & $\times$     & $\surd$                   & 96.9      \\
		Full model & $\surd$     & $\surd$                  & 98.9        \\
		\hline
	\end{tabular}
	}
	\vspace{0.2cm}
\end{table}

\begin{table}[!t]
\centering
   \caption{Generalization of PRN and MMF-GCN to other frameworks in terms of ADD-S (\%) and $<$2cm (\%) on YCB-Video.}
   \resizebox{0.45\textwidth}{!}{%
   \label{tab:ablation2}
   \renewcommand\arraystretch{1.2}
   \begin{tabular}{p{2.2cm}<{\centering}|p{1.1cm}<{\centering}|p{1cm}<{\centering}|p{1.1cm}<{\centering}|p{1.0cm}<{\centering}}
   \hline
          Method  & \multicolumn{2}{c|}{PVN3D} & \multicolumn{2}{c}{DenseFusion} \\ \hline
          Metric  & ADD-S        & \textless{}2cm       & ADD-S       & \textless{}2cm        \\  \hline
   Original model & 95.5 & 97.6 & 93.1 & 96.8           \\ 
   %\hline
   w/ PRN      & -            & -           & 94.1    & 97.2     \\ 
   %\hline
   w/ MMF-GCN & 96.2 & 98.4 & 93.5 &   97.2    \\ %\hline
   w/ both  & -            & -           &   94.9      & 98.1        \\ 
   \hline
   \end{tabular}
   }
%   \vspace{0.2cm}
   \end{table}

\begin{table*}[!t]
   \centering
      \caption{The influence of segmentation on different frameworks on YCB-Video in terms of AUC (\%) and $<$2cm (\%) (`-' indicates that the result is not reported).}
   \label{tab:ablation-segmentation}
   \resizebox{0.85\textwidth}{!}{%
   \begin{tabular}{c|c|c|c|c|c|c|c|c|c}
   \hline
   \multirow{2}{*}{} & \multicolumn{3}{c|}{PoseCNN   segmentation} & \multicolumn{3}{c|}{PVN3D segmentation} & \multicolumn{3}{c}{GT segmentation} \\ \cline{2-10} 
                     & PoseCNN      & DenseFusion      & Ours      & DenseFusion      & PVN3D     & Ours     & Densefusion     & PVN3D    & Ours    \\ \hline
   AUC               & 93.0         & 93.1             & \textbf{95.0}      & 91.8             & 95.5      & \textbf{95.8}     & 94.5            & 96.4     & \textbf{96.9}    \\ 
   %\hline
   \textless{}2cm              & 93.2         & 96.8             & \textbf{97.6}     & 92.8             & 97.6      & \textbf{98.5}     & 98.1            & -        & \textbf{99.9}   \\ \hline
   \end{tabular}
   %\vspace{-0.3cm}
   }
   \end{table*}
   
\begin{table}[!t]
\centering
\caption{Ablation study of the multi-resolution loss on YCB-Video in terms of ADD-S (\%) and $<$2cm (\%) .}
\resizebox{0.35\textwidth}{!}{%
\begin{tabular}{l|c|c|c}
\hline
      & WO-PRN & PRN-SR & PRN-MR \\ \hline
ADD-S & 94.0   &   94.6     & \textbf{95.8}   \\ 
\textless{} 2cm & 97.1   & 96.6  & \textbf{98.5} \\ \hline
\end{tabular}
}
\label{tab:PRN}
%\vspace{-0.1cm}
\end{table}

%It is only inferior to PVN3D. But the comparison is not fair since they employ multiple models in testing. Compare to the other counterparts, PR-GCN reaches the highest performance, demonstrating its effectiveness.

%-------------------------------------------------------------------------
   
%------------------------------------------------------------------------

\subsection{Ablation Study}

We comprehensively validate individual components of PR-GCN in the following.
 
\textbf{The impact of PRN and MMF-GCN.} The baseline method removes PRN and replaces MMF-GCN by DGCNN \cite{DGCNN} which adopts the same basic point cloud aggregator as our PR-GCN. As Table \ref{tab:ablation1} displays, PRN boosts the baseline by 2.0\%, indicating that refined point clouds contribute to pose estimation, while MMF-GCN achieves an improvement of 2.1\%, demonstrating its advantage in integrating multi-modal features. The combination of PRN and MMF-GCN further enhances the performance.

\textbf{The generalizability of PRN and MMF-GCN.} We generalize the PRN and MMF-GCN modules to two state-of-the-art frameworks including PVN3D \cite{pvn3d} and DenseFusion \cite{densefusion}, and evaluate their performance on YCB-Video. Note that PVN3D cannot utilize PRN directly, since segmentation is required on the whole scene while PRN focuses on specific objects. We thus only evaluate the effect of MMF-GCN on PVN3D. As shown in Table \ref{tab:ablation2}, PRN promotes the ADD-S of DenseFusion by 1\%, and a similar improvement can be observed when applying MMF-GCN. The results indicate that PRN and MMF-GCN benefit other frameworks for 6D pose estimation.
 
\textbf{The influence of segmentation.} As in Fig.~\ref{fig:pipeline}, our framework introduces RGB-based segmentation to extract foreground objects, while PoseCNN \cite{POSECNN}, DenseFusion \cite{densefusion} and PVN3D \cite{pvn3d} adopt different instance segmentation models. To validate the effect of segmentation, we replace the segmentation model in our framework by the counterparts used in PoseCNN and PVN3D as well as the ground-truth, and report their AUC and ADD-S$<$2cm metrics on YCB-Video. Similarly, we evaluate this factor on other frameworks, including PoseCNN, DenseFusion and PVN3D. As reported in Table~\ref{tab:ablation-segmentation}, all these frameworks achieve the highest AUC and ADD-S$<$2cm using ground-truth, indicating that better segmentation boosts the estimation accuracy. Meanwhile, with segmentation alternatives, our framework consistently outperforms the others, showing that PR-GCN is superior, regardless of which segmentation model is used.  

\begin{figure}[!t]
	\centering
	\includegraphics[width=0.98\linewidth]{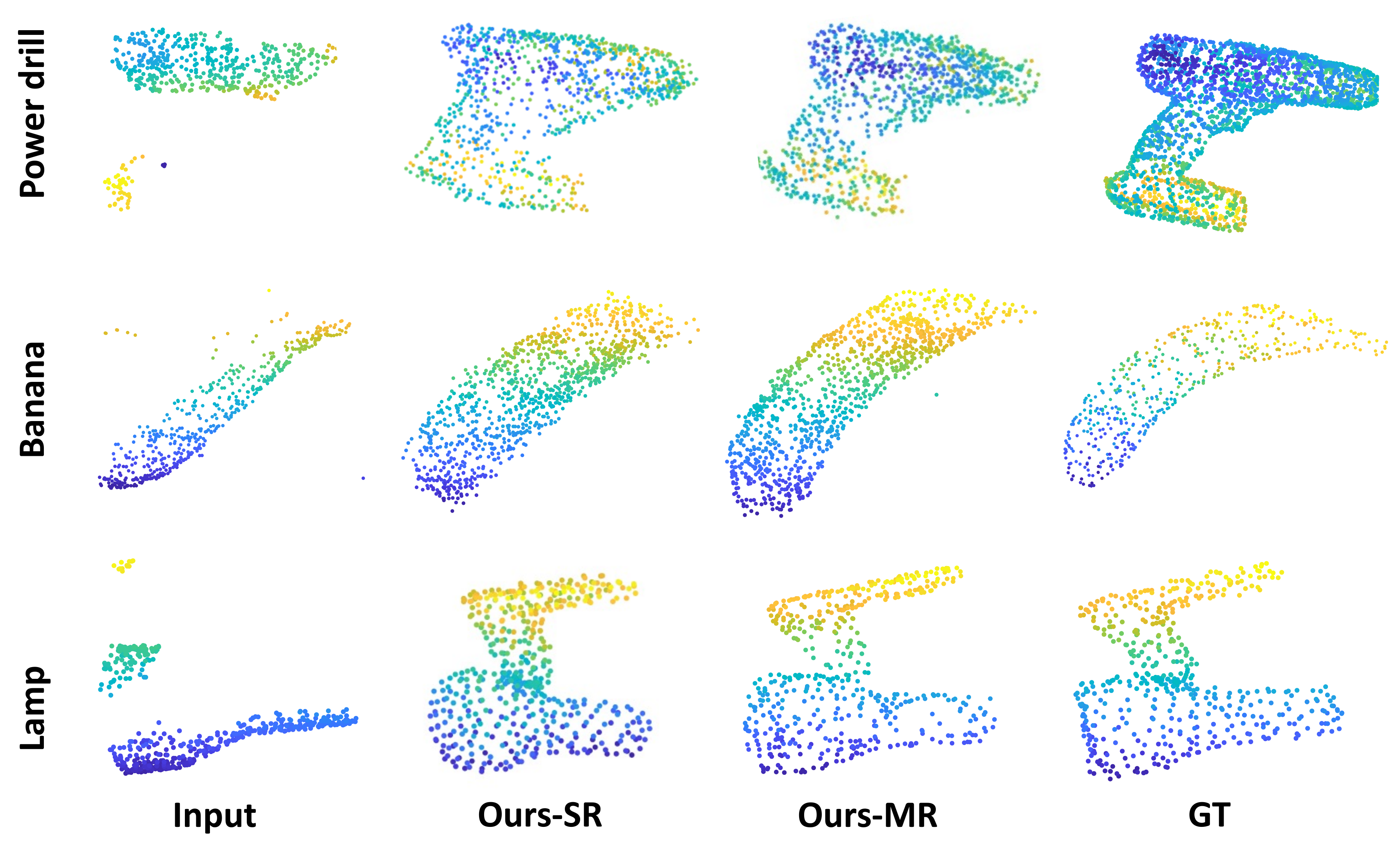}
	\caption{Visualization of the refined 3D points generated by PRN with/without the multi-resolution regression loss.}
	\label{fig:point_cloud}
%	\vspace{-0.2cm}
\end{figure} 

\textbf{The effect of the multi-resolution regression loss on PRN.} 
We finally validate the credit of the regularized multi-resolution regression loss $\mathcal{L}_{\rm{prn}}$. For comparison, we apply $\mathcal{L}_{\rm{prn}}$ on $\hat{\bm{P}}_{o}$ only, denoted by PRN-SR, while the multi-resolution case is denoted by PRN-MR. We also report the result without using PRN (WO-PRN). As summarized in Table \ref{tab:PRN}, adopting the single-resolution loss promotes the performance of our method. When the multi-resolution loss is added, ADD-S is further boosted to 95.8\%, demonstrating its effectiveness. Moreover, we visualize the generated refined 3D points in Fig.~\ref{fig:point_cloud}, and the results clearly show the advantage of PRN in dealing with the incompleteness and noise, after adding the loss $\mathcal{L}_{\rm{prn}}$.

%\begin{figure}[t]
%%\vspace{-0.05cm}
%\begin{center}
%    \includegraphics[width=0.98\linewidth]{failure.png}
%\end{center}
%   \caption{Failure cases of our method. Left: heavy occlusion (5\% visible) (`duck'). Right: symmetry object (`bowl').}
%\label{fig:failure}
%\vspace{-0.5cm}
%\end{figure}

%------------------------------------------------------------------------

\section{Conclusion}
In this paper, we propose a novel approach, namely deep Graph Convolutional Networks with Point Refinement (PR-GCN), to RGB-D based 6D pose estimation. We develop a Point Refinement Network (PRN) to improve the quality of depth representation, together with a Multi-Modal Fusion Graph Convolution Network (MMF-GCN) to fully explore local geometry-aware inter-modality correlations for sufficient combination. Extensive experiments validate its superiority and the PRN and MMF-GCN modules.

\section*{Acknowledgment}
This work is partly supported by the National Natural Science Foundation of China (No. 62022011), the Research Program of State Key Laboratory of Software Development Environment (SKLSDE-2021ZX-04), and the Fundamental Research Funds for the Central Universities.

{\small
\bibliographystyle{ieee_fullname}

}

\end{document}